\documentclass[pdflatex,sn-mathphys-num]{sn-jnl}% Math and Physical Sciences Numbered Reference Style
\usepackage{graphicx}%
\usepackage{multirow}%
\usepackage{amsmath,amssymb,amsfonts}%
\usepackage{amsthm}%
\usepackage{mathrsfs}%
\usepackage[title]{appendix}%
\usepackage{xcolor}%
\usepackage{textcomp}%
\usepackage{manyfoot}%
\usepackage{booktabs}%
\usepackage{algorithm}%
\usepackage{algorithmicx}%
\usepackage{algpseudocode}%
\usepackage{listings}%
\usepackage{svg}
\theoremstyle{thmstyleone}%
\theoremstyle{thmstyletwo}%

\theoremstyle{thmstylethree}%

\usepackage{booktabs}
\usepackage{pifont}

\newcommand{\cmark}{\ding{51}}
\newcommand{\xmark}{\ding{55}}

\begin{document}

\title[LAYA]{LAYA: Layer-wise Attention Aggregation for Interpretable Depth-Aware Neural Networks}


\author[]{\fnm{Gennaro} \sur{Vessio}}\email{gennaro.vessio@uniba.it}

\affil[]{\orgdiv{Department of Computer Science}, \orgname{University of Bari Aldo Moro}, \orgaddress{\city{Bari}, \country{Italy}}}


\abstract{Deep neural networks typically rely on the representation produced by their final hidden layer to make predictions, implicitly assuming that this single vector fully captures the semantics encoded across all preceding transformations. However, intermediate layers contain rich and complementary information—ranging from low-level patterns to high-level abstractions—that is often discarded when the decision head depends solely on the last representation. This paper revisits the role of the output layer and introduces \textbf{LAYA} (\textit{Layer-wise Attention Aggregator}), a novel output head that dynamically aggregates internal representations through attention. Instead of projecting only the deepest embedding, LAYA learns input-conditioned attention weights over layer-wise features, yielding an interpretable and architecture-agnostic mechanism for synthesizing predictions. Beyond improving feature aggregation, the learned attention coefficients provide intrinsic layer-attribution scores that explicitly quantify the contribution of each representation to the final decision, without requiring external post-hoc explanation methods. Experiments on image classification datasets show that LAYA achieves competitive predictive performance while producing meaningful depth-aware explanations. Furthermore, quantitative and qualitative interpretability analyses demonstrate that LAYA's attention scores closely reflect the actual contribution of individual layers, revealing structured, task-dependent patterns of depth utilization while providing intuitive explanations of how different abstraction levels contribute to each prediction.}




\maketitle

\section{Introduction}

The remarkable success of deep neural networks in computer vision is largely attributable to their ability to learn hierarchical representations, progressively transforming raw visual inputs into increasingly abstract representations. Yet, despite differences in architecture and learning paradigm, most architectures ultimately derive their predictions exclusively from the representation produced by the \emph{last hidden layer}. Formally, given an input $x$ processed by a model with parameters $\theta$, the forward pass produces a sequence of $L$ hidden representations
\[
h_1, h_2, \dots, h_L = f_\theta(x),
\]
where $f_\theta$ denotes the composition of the hidden layers. The final prediction is then obtained by applying a task-specific output head to the last representation:
\[
\hat{y} = \phi\big(W h_L + b\big),
\]
where $\phi(\cdot)$ denotes a task-specific activation function. This conventional design implicitly assumes that the last embedding $h_L$ encapsulates all relevant semantic information extracted by the network. Moreover, it provides little insight into how information encoded at different depths contributes to the final decision, making the role of intermediate representations difficult to interpret.

However, empirical and theoretical studies suggest that different layers capture distinct and complementary aspects of the input: early layers encode local patterns, middle layers capture structural relations, and deeper layers represent increasingly abstract semantics (e.g., \citep{bau2017network,tenney2019bert}). Popular architectures such as ResNet~\citep{he2016deep} and DenseNet~\citep{huang2017densely} partially address this issue through skip or dense connections, facilitating gradient flow and feature reuse. Nevertheless, these mechanisms operate within the feature extractor, while the output head remains a static projection over the final representation. Similarly, multi-scale fusion models such as FPN~\citep{lin2017feature} and Transformer-based architectures~\citep{vaswani2017attention,dosovitskiy2020image} exploit hierarchical information during feature learning but still collapse the resulting representational hierarchy before classification.

This work revisits the role of the output stage of deep neural networks. The central idea is that the output layer should not depend exclusively on the deepest representation, but should instead integrate information across depth while simultaneously exposing how different abstraction levels contribute to the final prediction. To this end, \textbf{LAYA} (\textit{Layer-wise Attention Aggregator}) is introduced as a lightweight output module that learns to weight and combine all hidden representations $\{h_i\}_{i=1}^L$ in an input-dependent way.

Each hidden state is mapped into a shared latent space through a lightweight adapter $g_i(\cdot)$, ensuring comparability across layers. LAYA then replaces the standard classifier head with an attention-based aggregator:
\[
h_{\text{agg}} = \sum_{i=1}^{L} \alpha_i(x)\, g_i(h_i),
\]
where $\alpha_i(x)$ are learnable, input-conditioned coefficients reflecting the relevance of each layer for the current sample. The final prediction is computed as
\[
\hat{y} = \phi(W h_{\text{agg}} + b).
\]

This formulation transforms the output layer from a passive projection into an active module that reasons over the network's internal representations. Crucially, the attention weights $\alpha_i(x)$ provide explicit layer-attribution scores, revealing how different depths contribute to each prediction. Unlike conventional post-hoc explanation methods, these relevance scores are generated as part of the forward pass itself, making interpretability an \textit{intrinsic} property of the model. Furthermore, they can be assessed both quantitatively, through faithfulness-based analyses, and qualitatively, by examining task-dependent attention patterns across network depth.

In summary, this work makes three main contributions. First, it highlights a fundamental limitation of conventional output heads, which rely exclusively on the deepest representation and therefore disregard both the complementary information encoded in intermediate layers and its interpretability. Second, it introduces LAYA, a lightweight, architecture-agnostic output module that performs input-conditioned aggregation across network depth while naturally producing intrinsic layer-attribution scores. Third, it presents a comprehensive experimental evaluation on different image classification datasets, combining predictive performance with quantitative and qualitative interpretability analyses to validate the fidelity and usefulness of the learned attention distributions.

The rest of the paper is structured as follows. Section~\ref{sec:related-work} surveys related work. Section~\ref{sec:method} presents the LAYA mechanism and its integration into standard architectures. Section~\ref{sec:experiments} describes the experimental protocol, reports the predictive performance, and presents the interpretability analyses. Finally, Section~\ref{sec:conclusion} concludes the paper and outlines directions for future research.

\section{Related Work}\label{sec:related-work}

\subsection{Hierarchical Representation Learning}

Deep neural networks organize computation hierarchically, with early layers capturing low-level features and deeper ones encoding progressively abstract semantics. Early convolutional models such as LeNet~\citep{lecun2002gradient}, AlexNet~\citep{krizhevsky2012imagenet}, and VGG~\citep{simonyan2014very} demonstrated the effectiveness of deep hierarchies but suffered from vanishing gradients and feature degradation. Residual Networks (ResNet)~\citep{he2016deep} introduced skip connections to facilitate gradient flow and feature reuse, enabling the training of substantially deeper architectures. Densely Connected Networks (DenseNet)~\citep{huang2017densely} further extended this principle by concatenating outputs from all preceding layers to encourage feature reuse and reduce redundancy. While these designs improve representational richness and allow information to flow across layers, the final classifier still receives a \emph{single, collapsed representation} $h_L$. Although $h_L$ is a function of all preceding transformations, the output layer cannot explicitly access intermediate representations nor modulate their contribution based on the input; all depth-specific information is ultimately compressed into the last embedding.

\subsection{Multi-scale and Feature Fusion Models}

Many architectures explicitly integrate features across multiple spatial scales or abstraction levels. Feature Pyramid Networks (FPN)~\citep{lin2017feature} and BiFPN~\citep{tan2020efficientdet} merge hierarchical features to enhance object detection and segmentation, while HRNet~\citep{sun2019deep}, U-Net++~\citep{zhou2018unet++}, and DeepLab~\citep{chen2018encoder} leverage encoder--decoder symmetries to preserve both spatial resolution and semantic depth. Vision Transformers (ViT)~\citep{dosovitskiy2020image} and hierarchical variants such as Swin Transformer~\citep{liu2021swin} similarly exploit multi-stage representations. CrossViT~\citep{chen2021crossvit} extends this idea through cross-attention between representations extracted at different patch granularities, while cross-layer attention mechanisms~\citep{huang2022cross} further demonstrate that intermediate representations contain complementary task-relevant information. More recently, LayerFusion~\citep{zheng2024layer} showed that aggregating intermediate representations can improve contextual embeddings, although such aggregation remains embedded within the feature extractor. Consequently, these approaches enhance representation learning but still rely on a conventional output head operating on a single final representation.

\subsection{Output-level Reasoning}

Only a limited number of approaches revisit the role of the prediction head itself. Early-exit architectures such as BranchyNet~\citep{teerapittayanon2016branchynet} introduce auxiliary classifiers at intermediate layers to improve inference efficiency, while deep supervision methods~\citep{lee2015deeply} propagate training signals throughout the network. However, neither family combines intermediate representations at inference time. Ensemble approaches~\citep{lakshminarayanan2017simple} aggregate predictions from multiple models but require multiple forward passes. Among existing methods, ScalarMix~\citep{peters2018deep} is conceptually the closest to our work, as it explicitly combines representations extracted from different depths. Nevertheless, its mixing coefficients are global parameters shared across all inputs, preventing the model from adapting the contribution of individual layers to each sample. In contrast, attention mechanisms~\citep{bahdanau2014neural} learn dynamic relevance scores, although they are typically applied across sequence elements rather than across network depth.

\subsection{Interpretability in Deep Neural Networks}

The increasing adoption of deep neural networks has stimulated extensive research on explainability and interpretability. Most existing methods are \emph{post-hoc}, producing explanations after the prediction has been computed without modifying the underlying model. Gradient-based attribution methods constitute one of the most widely adopted families of post-hoc explainability techniques. Representative examples include Integrated Gradients~\citep{sundararajan2017axiomatic}, Grad-CAM~\citep{selvaraju2017grad}, and Gradient$\times$Activation~\citep{ancona2018towards}, which estimate feature relevance by exploiting gradients and internal activations. More recently, attention-guided visualization methods have been proposed specifically for Vision Transformers, combining self-attention with gradient information to improve the faithfulness of visual explanations \citep{leem2024attention}. Other model-agnostic approaches, including LIME~\citep{ribeiro2016should} and SHAP~\citep{lundberg2017unified}, explain predictions through local surrogate models or feature attribution values. While these methods have become standard tools for explaining neural networks, they generally require additional computations beyond inference and provide explanations that are external to the prediction mechanism itself. By contrast, intrinsically interpretable models aim to expose their decision process directly through their internal computations, reducing the gap between prediction and explanation.

\subsection{Positioning of LAYA}

LAYA belongs to this latter family by combining output-level aggregation with intrinsic interpretability. Rather than relying exclusively on the deepest representation, it assigns \emph{input-conditioned} attention weights to representations extracted from all hidden layers and aggregates them before classification. The resulting attention coefficients naturally serve as layer-attribution scores, explicitly indicating how much each depth contributes to the prediction. Unlike ScalarMix, which learns a single global combination shared across all inputs, LAYA computes a different layer weighting for every sample. Unlike cross-layer fusion approaches, it leaves the backbone completely unchanged and operates exclusively as a lightweight output head. Finally, unlike post-hoc explanation methods, LAYA produces explanations as a direct consequence of the prediction process itself, requiring neither additional explanation models nor extra inference steps. This combination of architecture-agnostic integration, dynamic depth aggregation, and intrinsic interpretability distinguishes LAYA from existing approaches. To further clarify the positioning of LAYA with respect to the most closely related approaches, Table~\ref{tab:related-comparison} compares representative methods in terms of layer aggregation, dynamic feature weighting, interpretability, and the need for backbone modifications.

\begin{table}[t]
\centering
\footnotesize
\caption{Comparison of representative approaches for layer aggregation and neural network interpretability.}
\label{tab:related-comparison}
\renewcommand{\arraystretch}{1.15}
\begin{tabular}{lccccc}
\toprule
\textbf{Method} & \textbf{Layer} & \textbf{Input} & \textbf{Intrinsic} & \textbf{Post-hoc} & \textbf{Backbone} \\
 & \textbf{aggregation} & \textbf{dependent} & \textbf{explanations} & \textbf{explanations} & \textbf{changes} \\
\midrule
Standard classifier            & \xmark & \xmark & \xmark & \xmark & \xmark \\
ScalarMix                      & \cmark & \xmark & \xmark & \xmark & \xmark \\
% LayerFusion                    & \cmark & \cmark & \xmark & \xmark & \cmark \\
Integrated Gradients           & \xmark & --     & \xmark & \cmark & \xmark \\
% Grad-CAM                       & \xmark & --     & \xmark & \cmark & \xmark \\
Gradient$\times$Activation     & \xmark & --     & \xmark & \cmark & \xmark \\
\midrule
\textbf{LAYA (ours)}           & \textbf{\cmark} & \textbf{\cmark} & \textbf{\cmark} & \textbf{\xmark} & \textbf{\xmark} \\
\bottomrule
\end{tabular}
\end{table}

\section{Method}\label{sec:method}

\subsection{Overview}

Deep neural networks learn hierarchical representations, with different layers capturing increasingly abstract information. However, the standard output head relies exclusively on the final embedding $h_L$, discarding potentially useful information encoded at earlier depths. LAYA (\textit{Layer-wise Attention Aggregator}) redefines the standard output head as a depth-aware aggregation module. Instead of relying solely on $h_L$, LAYA assigns input-dependent attention weights to all hidden representations $\{h_i\}_{i=1}^{L}$ and forms a weighted combination prior to prediction. This enables the model to adaptively emphasize different levels of abstraction depending on the input. At the same time, the learned attention distribution provides an explicit description of which representations are considered most relevant for each individual prediction.

\begin{figure}
    \centering
    \includegraphics[width=\textwidth]{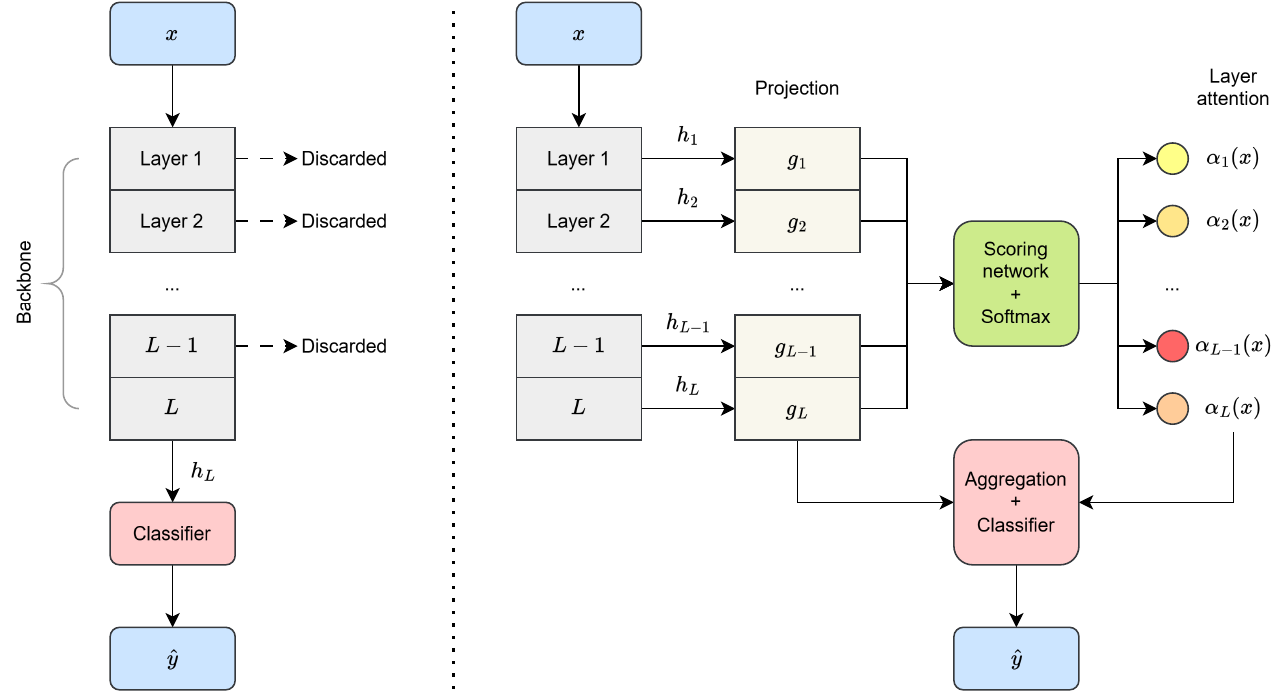}
    \caption{Comparison between a standard output head and LAYA. While the standard classifier relies exclusively on the final hidden representation $h_L$, LAYA projects and aggregates representations from all backbone layers through input-dependent attention weights. The learned coefficients $\alpha_i(x)$ jointly determine the aggregated representation used for prediction and provide intrinsic, sample-specific layer relevance scores. For visual clarity, the optional shared transformation $\psi(\cdot)$ applied before attention scoring is omitted.}
    \label{fig:laya-overview}
\end{figure}

Figure~\ref{fig:laya-overview} illustrates the main difference between a standard output head and LAYA.

% \begin{figure*}[t]
%     \centering
%     \includesvg[
%         width=\textwidth,
%         inkscapelatex=false
%     ]{LAYA}
%     \caption{Comparison between a standard output head and LAYA. While the standard classifier relies exclusively on the final hidden representation $h_L$, LAYA projects and aggregates representations from all backbone layers through input-dependent attention weights. The learned coefficients $\alpha_i(x)$ jointly determine the aggregated representation used for prediction and provide intrinsic, sample-specific layer relevance scores. For visual clarity, the optional shared transformation $\psi(\cdot)$ applied before attention scoring is omitted.}
%     \label{fig:laya-overview}
% \end{figure*}

\subsection{Formulation}

Consider a backbone network with $L$ hidden layers:
\[
h_i = f_i(h_{i-1}), \quad i = 1, \dots, L, \qquad h_0 = x.
\]

In standard architectures, the prediction is computed as:
\[
\hat{y} = \phi(W h_L + b).
\]

LAYA generalizes this formulation by treating all hidden representations $\{h_i\}$ as potentially informative. Each representation is first projected into a common latent space of dimension $d^\ast$ through a learnable adapter:
\[
g_i: \mathbb{R}^{d_i} \rightarrow \mathbb{R}^{d^\ast},
\qquad
z_i = g_i(h_i).
\]

An input-conditioned aggregation is then computed as
\[
h_{\mathrm{agg}} =
\sum_{i=1}^{L}
\alpha_i(x)\, z_i,
\]
where the coefficients $\alpha_i(x)$ represent attention weights specific to the input sample.

The final prediction becomes
\[
\hat{y}=\phi(W h_{\mathrm{agg}}+b).
\]

\subsection{Attention Mechanism}

Unlike conventional attention mechanisms that operate over spatial or temporal dimensions, LAYA performs attention over the representational depth of the network. The attention coefficients $\alpha_i(x)$ are generated by a lightweight meta-network operating on the enriched representations of all layers.

Each projected representation $z_i$ can optionally be transformed by a shared mapping
\[
u_i=\psi(z_i),
\]
where $\psi$ is either the identity function or a lightweight two-layer MLP. The identity mapping preserves a linear transformation of the enriched features, whereas the MLP introduces additional non-linearity. Both variants have been considered throughout the experimental evaluation.

The resulting representations $\{u_i\}$ are concatenated and processed by a scoring MLP that produces unnormalized relevance scores:
\[
s(x)=
\mathrm{MLP}_{\mathrm{score}}
([\,
u_1,
u_2,
\dots,
u_L
\,])
\in
\mathbb{R}^{L}.
\]

The scores are converted into attention coefficients through a temperature-scaled softmax:
\[
\alpha_i(x)=
\frac{\exp(s_i(x)/\tau)}
{\sum_{j=1}^{L}\exp(s_j(x)/\tau)},
\]
where $\tau>0$ controls the sharpness of the attention distribution, with lower values encouraging sparser and more selective allocations.

The aggregated representation is finally computed as
\[
h_{\mathrm{agg}}
=
\sum_{i=1}^{L}
\alpha_i(x)\,
g_i(h_i),
\]
and passed to the output classifier.

The attention coefficients $\alpha_i(x)$ serve as explicit, input-specific relevance scores over the hidden representations, directly indicating how much each layer contributes to the final prediction. Unlike post-hoc attribution methods, these scores are produced by the prediction mechanism itself and therefore constitute an intrinsic explanation of the model's decision process. Consequently, no additional explanation algorithm is required after inference. 

\subsection{Training and Integration}

LAYA is trained jointly with the backbone using the same optimization objective (e.g., cross-entropy), without requiring auxiliary losses or multi-stage training procedures. The additional computational overhead remains modest. The projection adapters introduce $O\!\left(\sum_{i=1}^{L} d_i d^\ast\right)$ parameters to map layer representations into the shared latent space, while the scoring network contributes at most $O(L d^{\ast 2})$ parameters to compute the attention weights.

Since LAYA operates entirely at the output stage, the backbone itself remains unchanged and the module can be used as a drop-in replacement for a standard output head. Unlike post-hoc explanation methods, which typically require an additional analysis step after prediction, LAYA produces layer-wise relevance scores intrinsically during inference, providing explanations without any dedicated explanation pipeline.

\section{Experiments}\label{sec:experiments}

The experimental evaluation was designed to assess both the predictive and interpretability properties of LAYA. Specifically, the experiments address three questions: (\textit{i}) whether adaptive layer aggregation improves or preserves predictive performance across different backbone architectures; (\textit{ii}) whether the learned attention coefficients provide faithful estimates of layer relevance when compared with established post-hoc attribution methods; and (\textit{iii}) whether the resulting attention patterns reveal meaningful, task-dependent strategies for exploiting the representational hierarchy.

\subsection{Experimental Setup}

\subsubsection{Datasets and Backbones}

The empirical evaluation considered three image classification settings designed to assess LAYA across distinct forms of representational hierarchy: a multilayer perceptron on Fashion-MNIST, a convolutional neural network on CIFAR-10, and a pretrained Vision Transformer on an artwork-style classification task. This selection made it possible to examine whether input-conditioned layer aggregation behaved differently when depth corresponded to successive fully connected transformations, increasingly abstract convolutional stages, or transformer encoder blocks.

Although relatively lightweight compared to modern large-scale benchmarks, these datasets were deliberately selected because they provide controlled yet representative evaluation settings. Since LAYA operates exclusively at the output stage, the objective was not to maximize absolute predictive performance, but to isolate the effect of the proposed aggregation mechanism under different representational hierarchies while minimizing confounding factors such as large-scale pretraining, extensive data augmentation, or highly optimized architectures.

Fashion-MNIST~\citep{xiao2017fashion} contains $60{,}000$ training and $10{,}000$ test grayscale images from 10 clothing categories at a resolution of $28{\times}28$ pixels. For this dataset, a three-layer MLP with hidden dimensions $[512,256,128]$ was employed. Each hidden layer was followed by LayerNorm~\citep{ba2016layer} and GELU activation~\citep{hendrycks2016gaussian}, and all three intermediate representations were exposed to the aggregation head.

CIFAR-10~\citep{krizhevsky2009learning} comprises $50{,}000$ training and $10{,}000$ test color images from 10 object categories at a resolution of $32{\times}32$ pixels. The adopted backbone was a lightweight CNN composed of three sequential depthwise--pointwise convolutional stages~\citep{howard2017mobilenets}, each followed by LayerNorm and GELU activation, with spatial downsampling in the later stages. Global average pooling was applied to the output of each stage to obtain the layer-wise representations used by LAYA.

The third setting was based on the \emph{Best Artworks of All Time} collection,\footnote{\url{https://www.kaggle.com/datasets/ikarus777/best-artworks-of-all-time}} which was reorganized as a multi-class artistic-style classification task. Only image data were used. The images were resized to $224{\times}224$ and processed using a pretrained ViT-Base/16 backbone~\citep{dosovitskiy2020image}, initialized from ImageNet-21k weights. The backbone was kept frozen, and the CLS-token representations produced by all transformer encoder blocks were exposed to the aggregation head. This setting provided a richer pretrained hierarchy in which different blocks could encode complementary perceptual and semantic properties, thereby enabling the analysis of layer-wise attribution in a more complex representation space.

\subsubsection{Output Heads}

To assess the contribution of the proposed aggregation strategy, four output heads were compared while keeping the backbone architecture and training protocol unchanged. The heads differed exclusively in how the layer-wise representations $\{h_i\}_{i=1}^{L}$ were combined prior to classification.

\begin{itemize}

\item \textsc{LastLayer}: Only the deepest representation was used for classification:
\[
\hat{y} = \phi(W h_L + b).
\]
This corresponds to the conventional output design adopted by most deep neural networks.

\item \textsc{Concat}: Each layer representation was first projected into a shared latent space through the corresponding adapter $g_i(\cdot)$. The projected representations were then concatenated and processed by a small MLP:
\[
h_{\mathrm{agg}}
=
\mathrm{MLP}_{\mathrm{post}}
\!\left(
[
g_1(h_1),
\dots,
g_L(h_L)
]
\right),
\qquad
\hat{y}
=
\phi(W h_{\mathrm{agg}} + b).
\]
Unlike attention-based aggregation, this strategy does not explicitly model the relative contribution of each layer.

\item \textsc{ScalarMix}~\citep{peters2018deep}: A set of learnable scalar coefficients determined the average contribution of each layer across the entire dataset:
\[
h_{\mathrm{agg}}
=
\sum_{i=1}^{L}
\alpha_i\,g_i(h_i),
\qquad
\alpha_i
=
\frac{e^{s_i}}
{\sum_{j=1}^{L}e^{s_j}},
\quad
s_i\in\mathbb{R}.
\]
Although \textsc{ScalarMix} was originally proposed to combine contextual representations in NLP, it can naturally be adapted as a generic output aggregation strategy, making it the most closely related baseline considered in this work. However, the aggregation weights are global model parameters and therefore remain identical for all input samples.

\item \textsc{LAYA}: LAYA extends the \textsc{ScalarMix} formulation by replacing the global aggregation coefficients with input-conditioned attention weights. Consequently, the contribution of each layer is dynamically adapted to the input, while simultaneously providing an intrinsic layer-wise explanation of the prediction process.

\end{itemize}

The selected baselines also provide a controlled decomposition of the main design choices underlying \textsc{LAYA}, thereby serving an ablation-like function.

\subsubsection{Interpretability Baselines}

To assess the quality of the intrinsic explanations produced by LAYA, four attribution strategies representing different explanation paradigms were considered. Let
\[
a(x)=\left[a_1(x),\ldots,a_L(x)\right]
\]
denote the vector of layer-wise attribution scores assigned to an input sample $x$, where $L$ is the number of network layers. 
For attribution methods producing unnormalized layer relevance scores, the resulting scores were normalized to sum to one, making them directly comparable to the probability distributions naturally produced by \textsc{ScalarMix} and \textsc{LAYA}.
The following attribution methods were evaluated:

\begin{itemize}

\item \textsc{Random}: A random normalized attribution vector was generated for each sample, providing a lower-bound reference corresponding to the absence of meaningful layer attribution.

\item \textsc{ScalarMix}: The learned global aggregation coefficients
\[
\boldsymbol{\alpha}=
[\alpha_1,\ldots,\alpha_L]
\]
were directly interpreted as layer-attribution scores. Unlike LAYA, as mentioned above, these coefficients are independent of the input and therefore assign the same layer importance to every sample.

\item \textsc{Integrated Gradients}~\citep{sundararajan2017axiomatic}.
Integrated Gradients were computed over the layer representations using a zero hidden-state baseline. The attribution assigned to layer $i$ was obtained by aggregating the feature-level integrated gradients,
\[
a_i(x)
=
\sum_{j=1}^{d}
\left|
(h_{ij}-h'_{ij})
\int_0^1
\frac{\partial p_{\hat c}\!\left(h'+\alpha(h-h')\right)}
{\partial h_{ij}}
\,d\alpha
\right|,
\]
where $h'$ denotes the baseline representation. The resulting layer scores were normalized before evaluation.

\item \textsc{Gradient$\times$Activation}: For each layer representation $h_i$, the attribution score was computed as
\[
a_i(x)
=
\sum_{j=1}^{d}
\left|
\frac{\partial p_{\hat c}(x)}
{\partial h_{ij}}
\,h_{ij}
\right|,
\]
where $p_{\hat c}(x)$ denotes the confidence of the predicted class and $d$ is the dimensionality of the layer representation. The resulting scores were subsequently normalized across layers.

\end{itemize}

% To ensure a fair comparison, all attribution methods were computed on the same trained LAYA predictor. Consequently, differences in the quantitative evaluation reflect only the attribution mechanism rather than changes in the underlying predictive model.

\subsubsection{Evaluation Metrics}\label{sec:evaluation-metrics}

Predictive performance was evaluated using classification accuracy and macro-F1. To quantitatively assess the quality of the layer-wise explanations, two complementary interpretability metrics were considered: \emph{Faithfulness} and \emph{Deletion@1}.

Let
\[
\Delta_i(x)
=
p_{\hat c}(x)
-
p_{\hat c}^{(-i)}(x)
\]
denote the contribution of layer $i$ for input $x$, estimated through a leave-one-layer-out (LOLO) perturbation analysis. Here, $p_{\hat c}(x)$ is the confidence assigned to the predicted class $\hat c$, whereas $p_{\hat c}^{(-i)}(x)$ denotes the confidence obtained after masking the representation of layer $i$ while leaving the remainder of the trained model unchanged.

Faithfulness measures the agreement between the attribution scores produced by an explanation method and the actual contribution of each layer. Specifically, for each sample, the Spearman rank correlation was computed as
\[
\rho(x)
=
\mathrm{Spearman}
\left(
a(x),
\Delta(x)
\right),
\]
where
\[
\Delta(x)=
[\Delta_1(x),\ldots,\Delta_L(x)].
\]
The final faithfulness score corresponds to the mean sample-wise correlation over the evaluation set. Faithfulness has become a widely adopted criterion for evaluating attribution methods, as it measures the extent to which an explanation reflects the actual decision process of the underlying model. Recent studies have also highlighted the challenges involved in reliably quantifying this property, particularly for Vision Transformers \citep{junyi2024faithfulness}.

Deletion@1 evaluates whether the layer receiving the highest attribution score is indeed the most influential for the prediction. Let
\[
i^\star
=
\arg\max_i a_i(x)
\]
be the layer identified as most relevant. The corresponding confidence drop is computed as
\[
D(x)
=
p_{\hat c}(x)
-
p_{\hat c}^{(-i^\star)}(x).
\]
Larger values indicate that the explanation method successfully identifies layers having a stronger influence on the final prediction. 

In addition, we report \textit{Positive Deletion}, defined as the fraction of samples for which removing the highest-ranked layer produces a positive confidence drop ($D(x)>0$). This metric measures how consistently the selected layer contributes positively to the predicted class across the evaluation set.

To ensure a fair comparison, all attribution methods were evaluated on the same trained LAYA predictor, so that differences in the reported scores reflect only the attribution mechanism rather than changes in the underlying predictive model.

\subsubsection{Training Details}

All experiments followed a consistent evaluation protocol across the considered benchmarks. For each dataset, the compared output heads (\textsc{LastLayer}, \textsc{Concat}, \textsc{ScalarMix}, and \textsc{LAYA}) shared exactly the same backbone, data split, optimization protocol, and random seeds, ensuring that any observed differences could be attributed exclusively to the output aggregation mechanism.

For Fashion-MNIST, models were trained with the Adam optimizer \citep{kingma2015adam}, learning rate $3\times10^{-4}$, and batch size $128$. Training lasted for at most $50$ epochs using a fixed $10\%$ validation split, with early stopping on validation accuracy and restoration of the best checkpoint.

For CIFAR-10, optimization was performed with AdamW \citep{loshchilov2019decoupled}, learning rate $10^{-3}$, weight decay $10^{-4}$, and batch size $128$. Models were trained for at most $80$ epochs using the same $10\%$ validation split and early stopping on validation accuracy.

For the artwork classification task, the pretrained ViT-Base/16 backbone initialized from ImageNet-21k weights was kept frozen, and only the output head was optimized. Images were resized to $224\times224$, normalized following the pretrained model, and split into stratified training, validation, and test sets (80\%/10\%/10\%). Optimization employed AdamW with learning rate $3\times10^{-4}$, no weight decay, batch size $128$, early stopping, and a maximum of $50$ epochs.

The LAYA hyperparameters were selected independently for each dataset by grid search. Fashion-MNIST explored $d^\ast\in\{64,96,128\}$, CIFAR-10 explored $d^\ast\in\{128,192,256\}$, whereas the artwork dataset considered $d^\ast\in\{128,256,512,768\}$. For all datasets, the temperature was searched over $\tau\in\{0.5,1.0,1.5\}$, the transformation $\psi$ over $\{\texttt{identity},\texttt{mlp}\}$, and the scorer width over $\{d^\ast,2d^\ast\}$. Each candidate configuration was evaluated over three random seeds, and the best one was selected according to the highest mean validation accuracy. The \textsc{Concat} and \textsc{ScalarMix} baselines underwent an analogous search over the projection dimension $d^\ast$ to ensure a fair comparison.

Following hyperparameter selection, all reported results were obtained from five independent runs using fixed random seeds. Classification accuracy and macro-F1 are reported as mean $\pm$ standard deviation across runs, together with two-sided $95\%$ confidence intervals computed using Student's $t$ distribution. For \textsc{LAYA}, the learned attention coefficients were additionally stored and used for the quantitative and qualitative interpretability analyses presented in the following sections.

\subsection{Experimental Results}

\subsubsection{Classification Performance}

Table~\ref{tab:results} reports the classification performance of the considered output heads on Fashion-MNIST, CIFAR-10, and Best Artworks. 
% Within each dataset, all methods shared the same backbone, data partitions, preprocessing pipeline, and optimization protocol; only the mechanism used to construct the final representation was varied. The reported values are averages over five independent runs.

On Fashion-MNIST, all output heads achieved comparable performance. \textsc{LAYA} obtained the highest mean accuracy, reaching $0.8834\pm0.0046$, compared with $0.8829\pm0.0063$ for \textsc{LastLayer}, $0.8807\pm0.0026$ for \textsc{ScalarMix}, and $0.8789\pm0.0020$ for \textsc{Concat}. The corresponding macro-F1 scores followed the same overall pattern, although the differences were small and the confidence intervals largely overlapped. 
% These results indicate that aggregating intermediate representations does not substantially alter performance on this relatively simple benchmark, where the final backbone representation is already highly discriminative.

A clearer advantage emerged on CIFAR-10. \textsc{LAYA} achieved an accuracy of $0.7212\pm0.0060$ and a macro-F1 score of $0.7205\pm0.0053$, obtaining the highest mean performance among the compared heads. \textsc{Concat} and \textsc{ScalarMix} reached accuracies of $0.7155\pm0.0096$ and $0.7144\pm0.0171$, respectively, whereas \textsc{LastLayer} obtained $0.7112\pm0.0236$. In addition to the higher mean values, \textsc{LAYA} exhibited the lowest variability across runs on this dataset. 
% This suggests that sample-dependent aggregation can be beneficial when classification relies on information distributed across multiple levels of the feature hierarchy.

The Best Artworks dataset produced a different ranking. \textsc{Concat} achieved the strongest overall performance, with an accuracy of $0.8167\pm0.0053$ and a macro-F1 score of $0.8022\pm0.0056$, followed by \textsc{ScalarMix}, which obtained $0.8081\pm0.0042$ accuracy and $0.8006\pm0.0010$ macro-F1. \textsc{LAYA}, instead, reached $0.7968\pm0.0178$ accuracy and $0.7712\pm0.0153$ macro-F1. Although it did not match the two strongest aggregation baselines, it substantially outperformed \textsc{LastLayer}, whose accuracy and macro-F1 were $0.6650\pm0.0108$ and $0.6421\pm0.0130$, respectively. 
% Thus, on the frozen Vision Transformer backbone, exploiting intermediate representations was considerably more effective than relying exclusively on its final hidden state.

Overall, the results show that no single aggregation strategy dominates under all experimental conditions. \textsc{LAYA} obtained the highest mean performance on Fashion-MNIST and CIFAR-10, while static concatenation was more effective on Best Artworks. Nevertheless, all three layer-aggregation approaches markedly improved over \textsc{LastLayer} on the artwork classification task. This finding is particularly relevant because it confirms that intermediate representations contain complementary information that may not be preserved completely in the final layer of a frozen pretrained backbone.

The predictive results should be interpreted as evidence that the proposed mechanism remains competitive with established aggregation strategies while providing an additional property that they do not offer: an intrinsic, input-dependent estimate of the contribution of each representation level. Accordingly, the main advantage of \textsc{LAYA} is not universal superiority in classification accuracy, but its ability to combine competitive predictive performance with directly inspectable layer-level attribution, as investigated in the following experiments.

\begin{table}[t]
\centering
\setlength{\tabcolsep}{6pt}
\renewcommand{\arraystretch}{1.10}
\resizebox{\linewidth}{!}{%
\begin{tabular}{l l c c c}
\toprule
\textbf{Dataset} & \textbf{Head} & \textbf{Accuracy} &
\textbf{Macro-F1} & \textbf{Acc.~95\% CI} \\
\midrule

\multirow{4}{*}{Fashion-MNIST}
& \textsc{LastLayer}
& $0.8829 \pm 0.0063$
& $0.8826 \pm 0.0066$
& $[0.8752,\,0.8907]$ \\

& \textsc{Concat}
& $0.8789 \pm 0.0020$
& $0.8783 \pm 0.0020$
& $[0.8764,\,0.8814]$ \\

& \textsc{ScalarMix}
& $0.8807 \pm 0.0026$
& $0.8800 \pm 0.0027$
& $[0.8775,\,0.8839]$ \\

& \textsc{LAYA}
& $0.8834 \pm 0.0046$
& $0.8827 \pm 0.0052$
& $[0.8777,\,0.8891]$ \\

\midrule

\multirow{4}{*}{CIFAR-10}
& \textsc{LastLayer}
& $0.7112 \pm 0.0236$
& $0.7108 \pm 0.0228$
& $[0.6819,\,0.7405]$ \\

& \textsc{Concat}
& $0.7155 \pm 0.0096$
& $0.7136 \pm 0.0117$
& $[0.7036,\,0.7274]$ \\

& \textsc{ScalarMix}
& $0.7144 \pm 0.0171$
& $0.7123 \pm 0.0177$
& $[0.6932,\,0.7356]$ \\

& \textsc{LAYA}
& $0.7212 \pm 0.0060$
& $0.7205 \pm 0.0053$
& $[0.7138,\,0.7286]$ \\

\midrule

\multirow{4}{*}{Best Artworks}
& \textsc{LastLayer}
& $0.6650 \pm 0.0108$
& $0.6421 \pm 0.0130$
& $[0.6516,\,0.6784]$ \\

& \textsc{Concat}
& $0.8167 \pm 0.0053$
& $0.8022 \pm 0.0056$
& $[0.8102,\,0.8233]$ \\

& \textsc{ScalarMix}
& $0.8081 \pm 0.0042$
& $0.8006 \pm 0.0010$
& $[0.8029,\,0.8134]$ \\

& \textsc{LAYA}
& $0.7968 \pm 0.0178$
& $0.7712 \pm 0.0153$
& $[0.7747,\,0.8189]$ \\

\bottomrule
\end{tabular}
}
\caption{Classification performance of the output heads across the three
datasets. Results are reported as mean $\pm$ standard deviation over five independent runs, together with the 95\% confidence interval for accuracy. Within each dataset, all methods shared the same backbone, data partitions, preprocessing pipeline, and training protocol.}
\label{tab:results}
\end{table}

\subsubsection{Quantitative Evaluation of Layer Attributions}
\label{sec:quantitative-interpretability}

Table~\ref{tab:interpretability-results} summarizes the quantitative comparison of the considered layer-attribution methods. All explanations were computed on the same trained \textsc{LAYA} predictor, ensuring that the observed differences originate exclusively from the attribution mechanism rather than from variations in the underlying predictive model. Following the protocol described in Section~\ref{sec:evaluation-metrics}, faithfulness was measured as the sample-wise Spearman correlation between the attribution ranking and the confidence variations produced by LOLO perturbations, whereas Deletion@1 measures the average confidence drop obtained after removing the layer identified as most relevant by each method. Higher values indicate more faithful explanations.

On Fashion-MNIST, the two gradient-based post-hoc explainers achieved the highest faithfulness scores, with mean Spearman correlations of $0.8957\pm0.2346$ for \textsc{Gradient$\times$Activation} and $0.8905\pm0.2337$ for \textsc{Integrated Gradients}. Nevertheless, the intrinsic attribution produced directly by \textsc{LAYA} remained remarkably close, reaching $0.8575\pm0.2896$, while substantially outperforming both the static \textsc{ScalarMix} coefficients ($0.2591\pm0.5683$) and the random baseline ($0.0028\pm0.7061$). The same trend is reflected by Deletion@1: removing the layer receiving the largest \textsc{LAYA} coefficient reduced the prediction confidence by $0.6946$ on average, compared with $0.7016$ for \textsc{Gradient$\times$Activation} and $0.6853$ for \textsc{Integrated Gradients}. Moreover, this removal produced a positive confidence drop for $98.98\%$ of the test samples, indicating that the layer receiving the highest intrinsic weight was almost always supportive of the final prediction.

\begin{table*}[t]
\centering
\setlength{\tabcolsep}{5pt}
\renewcommand{\arraystretch}{1.12}
\resizebox{\textwidth}{!}{%
\begin{tabular}{llccc}
\toprule
\textbf{Dataset} &
\textbf{Method} &
\textbf{Faithfulness ($\rho$)} &
\textbf{Deletion@1} &
\textbf{Positive Deletion} \\
\midrule

\multirow{5}{*}{Fashion-MNIST}
& \textsc{Random} & $0.0028 \pm 0.7061$ & $0.2509 \pm 0.3618$ & $0.8327$ \\
& \textsc{ScalarMix} & $0.2591 \pm 0.5683$ & $0.5501 \pm 0.3782$ & $0.9401$ \\
& \textsc{Integrated Gradients} & $0.8905 \pm 0.2337$ & $0.6853 \pm 0.2821$ & $0.9949$ \\
& \textsc{Gradient$\times$Activation} & $0.8957 \pm 0.2346$ & $0.7016 \pm 0.2587$ & $0.9941$ \\
& \textsc{LAYA} & $0.8575 \pm 0.2896$ & $0.6946 \pm 0.2719$ & $0.9898$ \\

\midrule

\multirow{5}{*}{CIFAR-10}
& \textsc{Random} & $0.0108 \pm 0.7071$ & $0.2347 \pm 0.3375$ & $0.8155$ \\
& \textsc{ScalarMix} & $0.9040 \pm 0.1989$ & $0.6715 \pm 0.2145$ & $0.9998$ \\
& \textsc{Integrated Gradients} & $0.9072 \pm 0.1964$ & $0.6715 \pm 0.2145$ & $0.9998$ \\
& \textsc{Gradient$\times$Activation} & $0.9041 \pm 0.1988$ & $0.6715 \pm 0.2145$ & $0.9998$ \\
& \textsc{LAYA} & $0.8299 \pm 0.2447$ & $0.6710 \pm 0.2153$ & $0.9997$ \\

\midrule

\multirow{5}{*}{Best Artworks}
& \textsc{Random} & $-0.0126 \pm 0.3001$ & $0.0594 \pm 0.1845$ & $0.6367$ \\
& \textsc{ScalarMix} & $0.3436 \pm 0.2694$ & $0.3372 \pm 0.3540$ & $0.9126$ \\
& \textsc{Integrated Gradients} & $0.4320 \pm 0.2600$ & $0.4736 \pm 0.3538$ & $0.9483$ \\
& \textsc{Gradient$\times$Activation} & $0.4247 \pm 0.2683$ & $0.5733 \pm 0.3193$ & $0.9581$ \\
& \textsc{LAYA} & $0.3627 \pm 0.2473$ & $0.5787 \pm 0.3257$ & $0.9569$ \\

\bottomrule
\end{tabular}
}
\caption{Quantitative comparison of layer-attribution methods. Faithfulness is measured as the sample-wise Spearman correlation between the attribution ranking and the confidence variations induced by leave-one-layer-out perturbations. Deletion@1 denotes the average confidence drop after removing the highest-ranked layer, while Positive Deletion reports the fraction of samples for which this removal decreased the predicted-class confidence. Higher values indicate more faithful attributions. Values are reported as mean $\pm$ standard deviation over all evaluated samples.}
\label{tab:interpretability-results}
\end{table*}

A similar picture emerged on CIFAR-10. The three post-hoc alternatives achieved very similar faithfulness values, with \textsc{Integrated Gradients}, \textsc{Gradient$\times$Activation}, and \textsc{ScalarMix} obtaining mean Spearman correlations of $0.9072$, $0.9041$, and $0.9040$, respectively. Although \textsc{LAYA} reached a slightly lower correlation ($0.8299\pm0.2447$), all learned methods produced virtually identical Deletion@1 scores (approximately $0.671$), and the confidence drop remained positive for more than $99.9\%$ of the evaluated samples. This suggests that the different attribution mechanisms generally identified the same dominant stage of the CNN, making Deletion@1 considerably less discriminative than the full LOLO rank correlation.

The Best Artworks benchmark proved substantially more challenging, resulting in lower faithfulness values for every method. Once again, the gradient-based approaches ranked first, with Spearman correlations of $0.4320\pm0.2600$ for \textsc{Integrated Gradients} and $0.4247\pm0.2683$ for \textsc{Gradient$\times$Activation}. The intrinsic \textsc{LAYA} coefficients achieved $0.3627\pm0.2473$, outperforming the static \textsc{ScalarMix} attribution ($0.3436\pm0.2694$) while remaining well above the random baseline ($-0.0126\pm0.3001$). Interestingly, \textsc{LAYA} obtained the highest Deletion@1 score on this benchmark ($0.5787$), slightly exceeding \textsc{Gradient$\times$Activation} ($0.5733$) and more clearly outperforming \textsc{Integrated Gradients} ($0.4736$). Therefore, although the gradient-based explainers produced rankings that more closely matched the exhaustive LOLO perturbation analysis, the layer assigned the largest intrinsic weight by \textsc{LAYA} was, on average, the most influential one for the final prediction.

Overall, the quantitative evaluation reveals a consistent pattern across the three datasets. As expected, the gradient-based post-hoc explainers achieved the highest faithfulness scores, since they explicitly exploit the local sensitivity of the trained predictor to produce explanations. However, \textsc{LAYA} consistently generated highly faithful intrinsic layer attributions without requiring any additional backward computation, remaining substantially more informative than both random attribution and, in most cases, static global weighting. Furthermore, its Deletion@1 performance remained competitive across all datasets and even represented the strongest result on Best Artworks. These findings indicate that the attention coefficients learned by \textsc{LAYA} provide meaningful estimates of layer relevance while being generated directly during the forward pass and simultaneously driving the prediction itself.

\subsubsection{Analysis of Learned Layer-Attention Patterns}

Beyond the quantitative interpretability evaluation, we analyzed the layer-wise attention patterns learned by \textsc{LAYA} to better understand how the model exploits the representational hierarchy across different datasets and classes.

\paragraph{Global Layer-wise Attention Profiles}

\begin{figure*}[t]
    \centering
    \includegraphics[width=\textwidth]{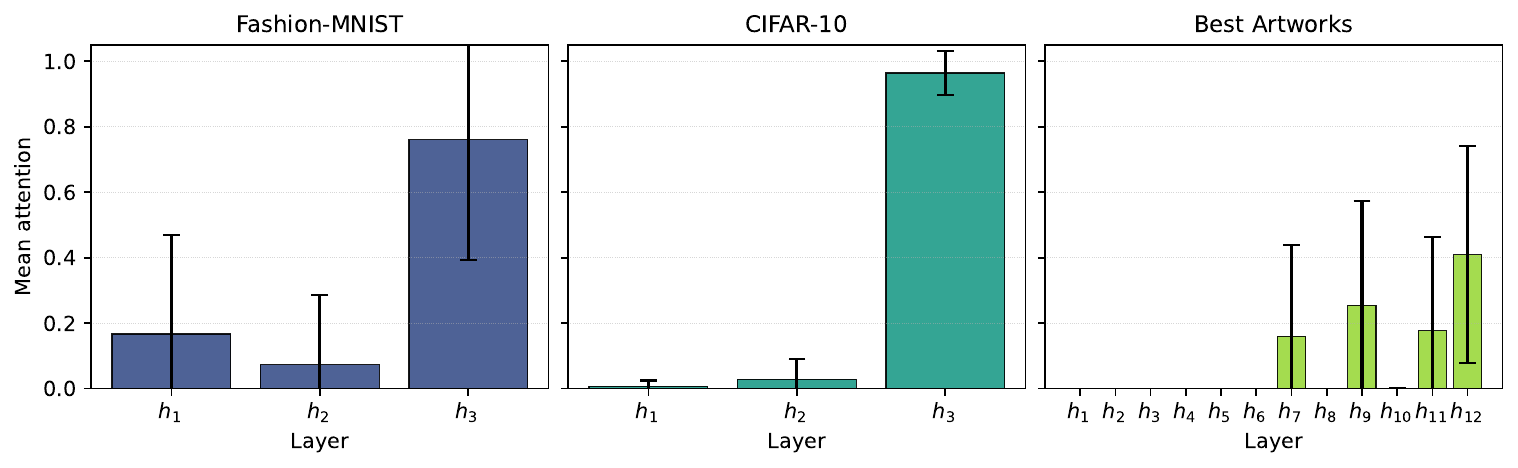}
    \caption{Global layer-wise attention profiles of \textsc{LAYA} on Fashion-MNIST, CIFAR-10, and Best Artworks. Bars and error bars report the mean and standard deviation of the attention coefficients over the full test set. The profiles reveal task-dependent depth usage: a predominantly deep but variable allocation on Fashion-MNIST, an almost complete concentration on the final stage on CIFAR-10, and a sparse allocation over intermediate and late ViT blocks on Best Artworks.}
    \label{fig:laya_global_attention}
\end{figure*}

To characterize the overall allocation of attention across depth, we computed the mean and standard deviation of $\alpha_i(x)$ over the full test set for each dataset. As shown in Fig.~\ref{fig:laya_global_attention}, the resulting profiles are strongly task-dependent. On Fashion-MNIST, attention is mainly concentrated on the deepest hidden layer, while the shallower representations retain non-negligible weights and substantial variability across samples. On CIFAR-10, the allocation becomes markedly sharper, with the final convolutional stage receiving almost all the attention. Best Artworks exhibits a different behavior: attention is sparse across the twelve ViT blocks and is concentrated on blocks $7$, $9$, $11$, and $12$, with the largest average weight assigned to the final block. The high variability of the active blocks indicates that different artworks can rely on different combinations of intermediate and late representations.

This multi-depth profile is compatible with the nature of artistic-style recognition, which may benefit from intermediate representations encoding texture, color organization, local geometry, and brushstroke-related cues, in addition to the higher-level information available in later blocks. Although the attention profiles alone do not establish which visual properties are encoded at each depth, they suggest that style classification does not rely exclusively on the most abstract representation.

% Overall, \textsc{LAYA} learns a flexible depth allocation on Fashion-MNIST, an almost exclusively deep strategy on CIFAR-10, and a sparse multi-depth strategy on Best Artworks.

\paragraph{Class-wise Attention Profiles}

\begin{figure*}[t]
    \centering
    \includegraphics[width=\textwidth]{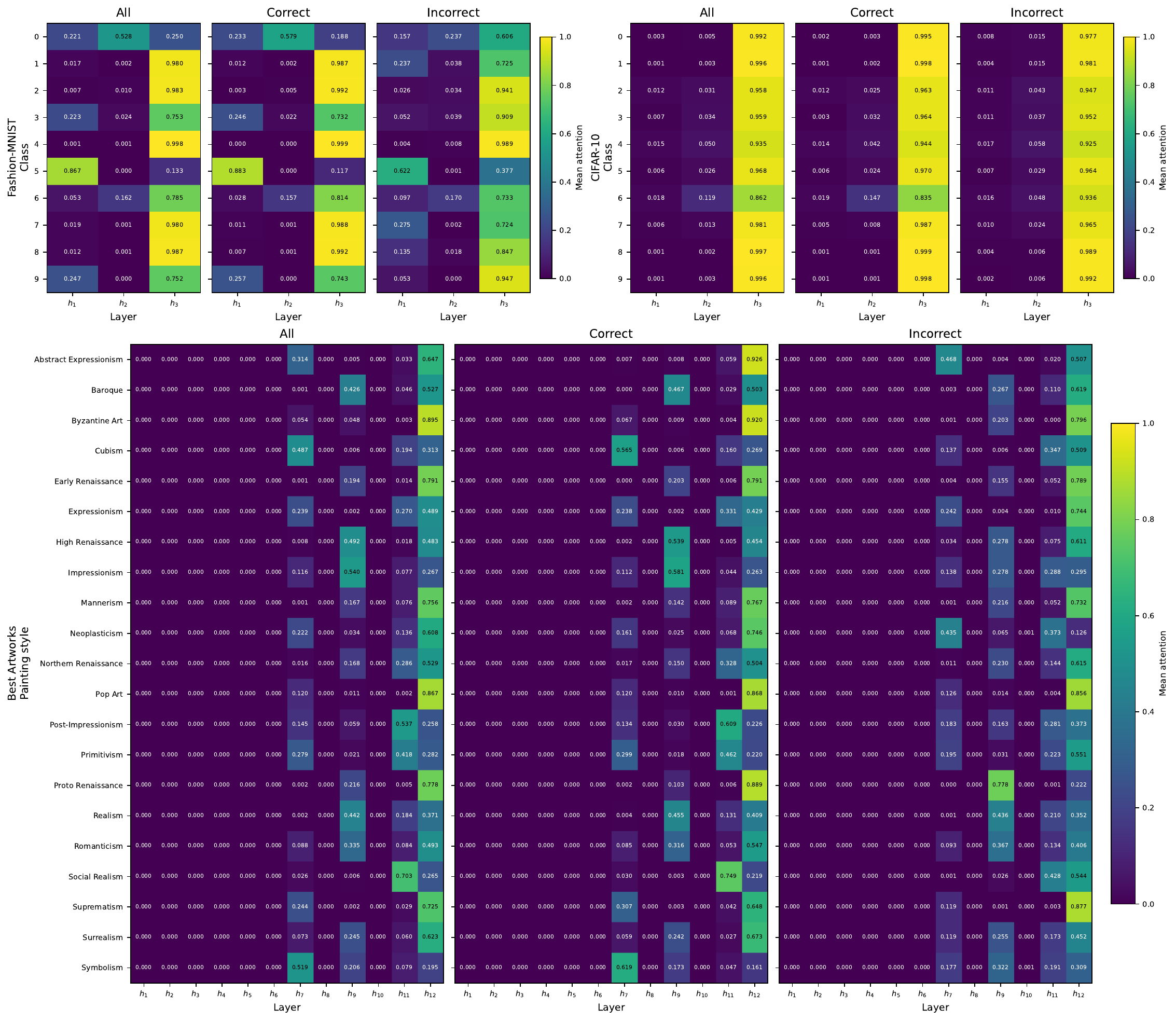}
    \caption{Class-wise mean attention profiles of \textsc{LAYA} on Fashion-MNIST, CIFAR-10, and Best Artworks. For each dataset, the heatmaps report all test samples, correctly classified samples, and incorrectly classified samples. Each cell contains the average attention coefficient $\bar{\alpha}_{c,i}$ assigned to layer $i$ for samples with true class $c$.}
    \label{fig:laya_classwise_attention}
\end{figure*}

To investigate class-dependent depth usage, attention coefficients were averaged for every class $c$ and layer $i$:
\[
\bar{\alpha}_{c,i}
=
\mathbb{E}\!\left[\alpha_i(x)\mid y=c\right].
\]
The analysis was further stratified into all, correctly classified, and incorrectly classified samples, as shown in Fig.~\ref{fig:laya_classwise_attention}.

On Fashion-MNIST, the profiles are markedly class-dependent. While most categories emphasize the deepest layer, some classes exhibit distinct strategies. Sandal images, for instance, primarily rely on the first layer, whereas T-shirt/top consistently assigns the largest weight to the intermediate layer, particularly among correctly classified samples. Classification errors are often accompanied by a redistribution of attention towards a different depth, suggesting that \textsc{LAYA} learns recurring class-specific profiles.

On CIFAR-10, the profiles are substantially more homogeneous: nearly all classes assign most of the attention to the deepest stage, independently of prediction correctness. This limited variability is consistent with the similar Deletion@1 scores observed across attribution methods, since different explainers tend to identify the same dominant representation.

Best Artworks displays the richest class-wise variation. Although block $12$ is globally dominant, different painting styles emphasize different subsets of blocks $7$, $9$, $11$, and $12$. Cubism and Symbolism rely primarily on block $7$, Impressionism on block $9$, and Post-Impressionism and Social Realism on block $11$, whereas several styles place most of their attention on block $12$. Correct and incorrect predictions also exhibit relevant shifts. For example, correct Abstract Expressionism predictions concentrate almost entirely on block $12$, whereas errors redistribute attention towards block $7$. Similar changes occur for Cubism, Neoplasticism, and Proto Renaissance. These patterns suggest that \textsc{LAYA} learns style-dependent depth profiles and that deviations from them can accompany classification errors.

Overall, the qualitative analysis complements the perturbation-based evaluation by revealing how \textsc{LAYA} exploits the representational hierarchy at both the global and class levels. The learned attention patterns highlight dataset-specific depth preferences and input-dependent variations, providing additional insight into the role of intermediate representations across different visual recognition tasks.

\section{Conclusion and Future Work}\label{sec:conclusion}

This work revisited the common design choice of relying exclusively on the final hidden representation for prediction in deep neural networks. The proposed LAYA demonstrates that the output stage can instead exploit the entire representational hierarchy by learning input-dependent attention weights across depth. Experiments on image classification datasets show that this simple modification provides competitive predictive performance while simultaneously providing intrinsic interpretability through layer-attribution scores that reveal how the model allocates attention across different levels of abstraction.

Beyond predictive performance, in fact, the learned attention profiles expose structured and task-dependent depth preferences that are directly available from the model, without requiring post-hoc explanation methods. These profiles provide a principled basis for \emph{attention-guided model compression}: layers that consistently receive negligible attention for a downstream task may represent promising candidates for simplification or removal, particularly when combined with fine-tuning or structural pruning.

Several additional research directions naturally emerge from this perspective. First, the observation that \textsc{LAYA} sometimes assigns substantial attention to intermediate representations suggests that future architectures could exploit these signals to support adaptive \emph{early-exit} mechanisms, enabling predictions once sufficient evidence has accumulated at earlier depths while providing an interpretable justification for the exit decision. Second, the observed differences between the attention profiles of correctly and incorrectly classified samples suggest potential diagnostic applications, including the identification of difficult or out-of-distribution examples, the analysis of failure modes, and support for dataset curation. Finally, the layer-attribution scores produced by \textsc{LAYA} offer new opportunities for studying how representations evolve across depth, providing a simple tool for investigating layer specialization and the relative importance of shallow versus deep abstractions across architectures and tasks.

Although this work focuses on computer vision, an important direction for future research is the extension of LAYA to transformer-based language models and other sequence architectures. Applying depth-aware aggregation to natural language processing could provide further insights into how linguistic information is progressively refined across transformer layers while preserving the intrinsic interpretability of the proposed layer-attention mechanism. More broadly, the layer-wise attribution scores produced by LAYA may offer a complementary perspective to emerging mechanistic interpretability approaches, which seek to understand the internal computations and representations learned by deep neural networks \citep{shu2025survey}. Integrating depth-aware aggregation with recent techniques for analyzing internal representations could provide a richer understanding of how information is progressively transformed across network depth. Extending the proposed framework to multimodal foundation models also represents a promising avenue for investigating how visual and linguistic representations are jointly exploited across layers.

\section*{Declaration of Generative AI and AI-Assisted Technologies in the Preparation of This Manuscript}

During the preparation of this manuscript, the author used ChatGPT (OpenAI) to assist with language refinement and improve the clarity of the presentation. All generated content was carefully reviewed, edited where necessary, and verified by the author, who takes full responsibility for the final version of the manuscript.

\section*{Acknowledgments}

The author would like to thank Giovanna Castellano for valuable discussions that helped clarify and refine key aspects of this work, as well as for her constructive feedback on the manuscript.


\bibliography{sn-bibliography}% common bib file

@article{lecun2002gradient,
	title        = {{Gradient-based learning applied to document recognition}},
	author       = {LeCun, Yann and Bottou, L{\'e}on and Bengio, Yoshua and Haffner, Patrick},
	year         = 2002,
	journal      = {Proceedings of the IEEE},
	publisher    = {Ieee},
	volume       = 86,
	number       = 11,
	pages        = {2278--2324}
}

@article{krizhevsky2012imagenet,
	title        = {{ImageNet classification with deep convolutional neural networks}},
	author       = {Krizhevsky, Alex and Sutskever, Ilya and Hinton, Geoffrey E},
	year         = 2012,
	journal      = {Advances in neural information processing systems},
	volume       = 25
}

@article{simonyan2014very,
	title        = {{Very deep convolutional networks for large-scale image recognition}},
	author       = {Simonyan, Karen and Zisserman, Andrew},
	year         = 2014,
	journal      = {arXiv preprint arXiv:1409.1556}
}

@inproceedings{he2016deep,
	title        = {{Deep residual learning for image recognition}},
	author       = {He, Kaiming and Zhang, Xiangyu and Ren, Shaoqing and Sun, Jian},
	year         = 2016,
	booktitle    = {Proceedings of the IEEE conference on computer vision and pattern recognition},
	pages        = {770--778}
}

@inproceedings{huang2017densely,
	title        = {{Densely connected convolutional networks}},
	author       = {Huang, Gao and Liu, Zhuang and Van Der Maaten, Laurens and Weinberger, Kilian Q},
	year         = 2017,
	booktitle    = {Proceedings of the IEEE conference on computer vision and pattern recognition},
	pages        = {4700--4708}
}

@inproceedings{lin2017feature,
	title        = {{Feature pyramid networks for object detection}},
	author       = {Lin, Tsung-Yi and Doll{\'a}r, Piotr and Girshick, Ross and He, Kaiming and Hariharan, Bharath and Belongie, Serge},
	year         = 2017,
	booktitle    = {Proceedings of the IEEE conference on computer vision and pattern recognition},
	pages        = {2117--2125}
}

@inproceedings{tan2020efficientdet,
	title        = {{EfficientDet: Scalable and efficient object detection}},
	author       = {Tan, Mingxing and Pang, Ruoming and Le, Quoc V},
	year         = 2020,
	booktitle    = {Proceedings of the IEEE/CVF conference on computer vision and pattern recognition},
	pages        = {10781--10790}
}

@inproceedings{sun2019deep,
	title        = {{Deep high-resolution representation learning for human pose estimation}},
	author       = {Sun, Ke and Xiao, Bin and Liu, Dong and Wang, Jingdong},
	year         = 2019,
	booktitle    = {Proceedings of the IEEE/CVF conference on computer vision and pattern recognition},
	pages        = {5693--5703}
}

@inproceedings{zhou2018unet++,
	title        = {{UNet++: A nested U-Net architecture for medical image segmentation}},
	author       = {Zhou, Zongwei and Rahman Siddiquee, Md Mahfuzur and Tajbakhsh, Nima and Liang, Jianming},
	year         = 2018,
	booktitle    = {International workshop on deep learning in medical image analysis},
	pages        = {3--11},
	organization = {Springer}
}

@inproceedings{chen2018encoder,
	title        = {{Encoder-decoder with atrous separable convolution for semantic image segmentation}},
	author       = {Chen, Liang-Chieh and Zhu, Yukun and Papandreou, George and Schroff, Florian and Adam, Hartwig},
	year         = 2018,
	booktitle    = {Proceedings of the European conference on computer vision (ECCV)},
	pages        = {801--818}
}

@article{vaswani2017attention,
	title        = {{Attention is all you need}},
	author       = {Vaswani, Ashish and Shazeer, Noam and Parmar, Niki and Uszkoreit, Jakob and Jones, Llion and Gomez, Aidan N and Kaiser, {\L}ukasz and Polosukhin, Illia},
	year         = 2017,
	journal      = {Advances in neural information processing systems},
	volume       = 30
}

@article{dosovitskiy2020image,
	title        = {{An image is worth 16x16 words: Transformers for image recognition at scale}},
	author       = {Dosovitskiy, Alexey},
	year         = 2020,
	journal      = {arXiv preprint arXiv:2010.11929}
}

@inproceedings{liu2021swin,
	title        = {{Swin transformer: Hierarchical vision transformer using shifted windows}},
	author       = {Liu, Ze and Lin, Yutong and Cao, Yue and Hu, Han and Wei, Yixuan and Zhang, Zheng and Lin, Stephen and Guo, Baining},
	year         = 2021,
	booktitle    = {Proceedings of the IEEE/CVF international conference on computer vision},
	pages        = {10012--10022}
}

@inproceedings{zheng2024layer,
	title        = {{Layer-wise representation fusion for compositional generalization}},
	author       = {Zheng, Yafang and Lin, Lei and Li, Shuangtao and Yuan, Yuxuan and Lai, Zhaohong and Liu, Shan and Fu, Biao and Chen, Yidong and Shi, Xiaodong},
	year         = 2024,
	booktitle    = {Proceedings of the AAAI Conference on Artificial Intelligence},
	volume       = 38,
	pages        = {19706--19714}
}

@inproceedings{teerapittayanon2016branchynet,
	title        = {{BranchyNet: Fast inference via early exiting from deep neural networks}},
	author       = {Teerapittayanon, Surat and McDanel, Bradley and Kung, Hsiang-Tsung},
	year         = 2016,
	booktitle    = {2016 23rd international conference on pattern recognition (ICPR)},
	pages        = {2464--2469},
	organization = {IEEE}
}

@article{huang2022cross,
	title        = {{Cross-layer attention network for fine-grained visual categorization}},
	author       = {Huang, Ranran and Wang, Yu and Yang, Huazhong},
	year         = 2022,
	journal      = {arXiv preprint arXiv:2210.08784}
}

@inproceedings{chen2021crossvit,
	title        = {{CrossViT: Cross-attention multi-scale vision transformer for image classification}},
	author       = {Chen, Chun-Fu Richard and Fan, Quanfu and Panda, Rameswar},
	year         = 2021,
	booktitle    = {Proceedings of the IEEE/CVF international conference on computer vision},
	pages        = {357--366}
}

@article{bahdanau2014neural,
	title        = {{Neural machine translation by jointly learning to align and translate}},
	author       = {Bahdanau, Dzmitry and Cho, Kyunghyun and Bengio, Yoshua},
	year         = 2014,
	journal      = {arXiv preprint arXiv:1409.0473}
}

@article{peters2018deep,
	title        = {{Deep contextualized word representations}},
	author       = {Peters, Matthew E. and Neumann, Mark and Iyyer, Mohit and Gardner, Matt and Clark, Christopher and Lee, Kenton and Zettlemoyer, Luke},
	year         = 2018,
    journal      = {arXiv preprint arXiv:1802.05365}
}

@article{lakshminarayanan2017simple,
	title        = {{Simple and scalable predictive uncertainty estimation using deep ensembles}},
	author       = {Lakshminarayanan, Balaji and Pritzel, Alexander and Blundell, Charles},
	year         = 2017,
	journal      = {Advances in neural information processing systems},
	volume       = 30
}

@inproceedings{lee2015deeply,
	title        = {{Deeply-supervised nets}},
	author       = {Lee, Chen-Yu and Xie, Saining and Gallagher, Patrick and Zhang, Zhengyou and Tu, Zhuowen},
	year         = 2015,
	booktitle    = {Artificial intelligence and statistics},
	pages        = {562--570},
	organization = {Pmlr}
}

@inproceedings{ribeiro2016should,
	title        = {{``Why should I trust you?'' Explaining the predictions of any classifier}},
	author       = {Ribeiro, Marco Tulio and Singh, Sameer and Guestrin, Carlos},
	year         = 2016,
	booktitle    = {Proceedings of the 22nd ACM SIGKDD international conference on knowledge discovery and data mining},
	pages        = {1135--1144}
}

@article{lundberg2017unified,
	title        = {{A unified approach to interpreting model predictions}},
	author       = {Lundberg, Scott M and Lee, Su-In},
	year         = 2017,
	journal      = {Advances in neural information processing systems},
	volume       = 30
}

@inproceedings{selvaraju2017grad,
	title        = {{Grad-CAM: Visual explanations from deep networks via gradient-based localization}},
	author       = {Selvaraju, Ramprasaath R and Cogswell, Michael and Das, Abhishek and Vedantam, Ramakrishna and Parikh, Devi and Batra, Dhruv},
	year         = 2017,
	booktitle    = {Proceedings of the IEEE international conference on computer vision},
	pages        = {618--626}
}

@article{tenney2019bert,
	title        = {{BERT rediscovers the classical NLP pipeline}},
	author       = {Tenney, Ian and Das, Dipanjan and Pavlick, Ellie},
	year         = 2019,
	journal      = {arXiv preprint arXiv:1905.05950}
}

@inproceedings{bau2017network,
	title        = {{Network dissection: Quantifying interpretability of deep visual representations}},
	author       = {Bau, David and Zhou, Bolei and Khosla, Aditya and Oliva, Aude and Torralba, Antonio},
	year         = 2017,
	booktitle    = {Proceedings of the IEEE conference on computer vision and pattern recognition},
	pages        = {6541--6549}
}

@article{xiao2017fashion,
	title        = {{Fashion-MNIST: A novel image dataset for benchmarking machine learning algorithms}},
	author       = {Xiao, Han and Rasul, Kashif and Vollgraf, Roland},
	year         = 2017,
	journal      = {arXiv preprint arXiv:1708.07747}
}

@article{krizhevsky2009learning,
	title        = {{Learning multiple layers of features from tiny images}},
	author       = {Krizhevsky, Alex and Hinton, Geoffrey},
	year         = 2009,
	publisher    = {Toronto, ON, Canada},
	journal      = {Technical Report, University of Toronto}
}

@article{ba2016layer,
	title        = {{Layer normalization}},
	author       = {Ba, Jimmy Lei and Kiros, Jamie Ryan and Hinton, Geoffrey E},
	year         = 2016,
	journal      = {arXiv preprint arXiv:1607.06450}
}

@article{hendrycks2016gaussian,
	title        = {{Gaussian Error Linear Units (GELUs)}},
	author       = {Hendrycks, D},
	year         = 2016,
	journal      = {arXiv preprint arXiv:1606.08415}
}

@article{howard2017mobilenets,
	title        = {{MobileNets: Efficient convolutional neural networks for mobile vision applications}},
	author       = {Howard, Andrew G and Zhu, Menglong and Chen, Bo and Kalenichenko, Dmitry and Wang, Weijun and Weyand, Tobias and Andreetto, Marco and Adam, Hartwig},
	year         = 2017,
	journal      = {arXiv preprint arXiv:1704.04861}
}

@inproceedings{kingma2015adam,
	title        = {{Adam: A Method for Stochastic Optimization}},
	author       = {Kingma, Diederik P and Ba, Jimmy},
	year         = 2015,
	booktitle    = {International conference on learning representations (ICLR)}
}

@inproceedings{sundararajan2017axiomatic,
  title={Axiomatic attribution for deep networks},
  author={Sundararajan, Mukund and Taly, Ankur and Yan, Qiqi},
  booktitle={International conference on machine learning},
  pages={3319--3328},
  year={2017},
  organization={PMLR}
}

@inproceedings{ancona2018towards,
  title={Towards better understanding of gradient-based attribution methods for Deep Neural Networks},
  author={Ancona, Marco and Ceolini, Enea and {\"O}ztireli, Cengiz and Gross, Markus},
  booktitle={International Conference on Learning Representations},
  year={2018}
}

@inproceedings{loshchilov2019decoupled,
  title={Decoupled Weight Decay Regularization},
  author={Loshchilov, Ilya and Hutter, Frank},
  booktitle={International Conference on Learning Representations},
  year={2019}
}

@article{shu2025survey,
  title={A survey on sparse autoencoders: Interpreting the internal mechanisms of large language models},
  author={Shu, Dong and Wu, Xuansheng and Zhao, Haiyan and Rai, Daking and Yao, Ziyu and Liu, Ninghao and Du, Mengnan},
  year={2025}
}

@inproceedings{junyi2024faithfulness,
  title={On the faithfulness of vision transformer explanations},
  author={Junyi, Wu and Weitai, Kang and Hao, Tang and Yuan, Hong and Yan, Yan},
  year={2024},
  organization={IEEE Conference on Computer Vision and Pattern Recognition}
}

@inproceedings{leem2024attention,
  title={Attention guided CAM: visual explanations of vision transformer guided by self-attention},
  author={Leem, Saebom and Seo, Hyunseok},
  booktitle={Proceedings of the AAAI conference on artificial intelligence},
  volume={38},
  number={4},
  pages={2956--2964},
  year={2024}
}

\end{document}